\def\year{2022}\relax
\documentclass[letterpaper]{article} 
\usepackage{aaai22}  
\usepackage{times}  
\usepackage{helvet}  
\usepackage{courier}  
\usepackage[hyphens]{url}  
\usepackage{graphicx} 
\usepackage{natbib}  
\usepackage{caption} 
\usepackage{subcaption}
\DeclareCaptionStyle{ruled}{labelfont=normalfont,labelsep=colon,strut=off} 
\usepackage{algpseudocode}
\usepackage{algorithm}
\usepackage{amsmath}
\usepackage{amssymb}

\usepackage{newfloat}
\usepackage{listings}
\floatstyle{ruled}
\newfloat{listing}{tb}{lst}{}
\floatname{listing}{Listing}
\title{Investigating Conversion from Mild Cognitive Impairment to Alzheimer's Disease using Latent Space Manipulation}
\author{
    Deniz Sezin Ayvaz \textsuperscript{\rm 1} and 
    \.{I}nci M. Bayta\c{s} \textsuperscript{\rm 1}
}
\affiliations{
    \textsuperscript{\rm 1}Department of Computer Engineering, Boğaziçi University\\

    Bebek, Istanbul, Turkey 34342\\
   sezin.ayvaz@boun.edu.tr, inci.baytas@boun.edu.tr
    
}

\usepackage{bibentry}
\nocopyright
\begin{document}

\maketitle

\begin{abstract}
Alzheimer’s disease is the most common cause of dementia that affects millions of lives worldwide. Investigating the underlying causes and risk factors of Alzheimer's disease is essential to prevent its progression. Mild Cognitive Impairment (MCI) is considered an intermediate stage before Alzheimer's disease. Early prediction of the conversion from the MCI to Alzheimer's is crucial to take necessary precautions for decelerating the progression and developing suitable treatments. In this study, we propose a deep learning framework to discover the variables which are identifiers of the conversion from MCI to Alzheimer's disease. In particular, the latent space of a variational auto-encoder network trained with the MCI and Alzheimer’s patients is manipulated to obtain the significant attributes and decipher their behavior that leads to the conversion from MCI to Alzheimer’s disease. By utilizing a generative decoder and the dimensions that lead to the Alzheimer's diagnosis, we generate synthetic dementia patients from MCI patients in the dataset. Experimental results show promising quantitative and qualitative results on one of the most extensive and commonly used Alzheimer’s disease neuroimaging datasets in literature.

\end{abstract}

\section{Introduction}
\label{sect:intro}
As the average lifespan increases, neurodegenerative diseases are becoming a common concern threatening millions of lives worldwide. Alzheimer's disease (AD) is one of the most well-known causes of dementia that affect cognitive functions~\cite{hou2019ageing}. According to the World Health Organization, there are approximately 50 million dementia patients in the world~\cite{who}. The number of AD patients in the world is expected to exceed 131 million by 2050~\cite{jiang2018correlation}. As the number of patients grows, researchers have been investigating the cause of AD. Fortunately, years of research shed light upon many unknowns. For instance, it is known that age, environment, and medications can trigger AD~\cite {hou2019ageing}. On the other hand, there are still many unknowns regarding the root cause and treatment of AD.

Dementia is a condition that impairs memory, language, mood, skills, recognition, judgment, and behavior such that the daily activities are degraded~\cite{cdc}. The transition between normal cognitive functions and dementia is called Mild Cognitive Impairment (MCI)~\cite{LIBON201472}.  Since the effect of the MCI on daily activities may not be noticeable, it is not classified as dementia~\cite{mckhann2011diagnosis}. While 1-2\% of the cognitively normal elderly develop dementia each year, approximately 12\% of those with MCI convert to dementia~\cite{plassman2008prevalence}. While some patients with MCI symptoms have AD in the later stages, some patients may remain in the MCI stage or progress very slowly~\cite{minhas2017predicting}. Therefore, the prediction of the conversion from MCI to AD is of great importance. 

AD can be diagnosed with an accuracy close to 95\%~\cite{mucke2009alzheimer}. However, for the prediction of early diagnosis, it is imperative to (i) take a detailed clinical history of the patient and their families, (ii) perform a neurological examination, (iii) perform neuropsychological tests appropriate for the education level to evaluate the cognitive functions of the patient, and (iv) evaluate the brain tissue with neuroimaging methods, such as Magnetic Resonance Imaging (MRI), and Positron Emission Tomography (PET). It is also essential to examine a large-scale patient cohort and simultaneously consider various variables. Artificial intelligence offers tools to simultaneously analyze multiple factors for a large patient cohort. In particular, machine learning and deep learning facilitate early diagnosis with predictive models. 

Although deep learning techniques are very successful, they are not easily interpretable. In the healthcare domain, it is crucial to know the underlying reasons behind the behavior of a model. To fully support the clinical research of dementia, predictive techniques should uncover the behavior of the input variables that leads to the transition between the stages of the disease. In this study, we propose a deep learning framework to discover the factors that lead to the conversion from MCI to AD. For this purpose, a latent space manipulation technique is adopted to obtain the significant attributes and decipher their behavior. The contributions of the proposed study are outlined below.
\begin{itemize}
\item A variational auto-encoder (VAE) network is trained to learn a latent representation for the patient data. The latent space of the encoder does not contain any label information.
\item A latent space manipulation technique is applied to the latent representations of the MCI patients. The generative decoder of the VAE outputs new patient data from the manipulated latent representations. To the best of our knowledge, the proposed study is the first attempt to apply latent space manipulation to tabular patient data. 
\item The manipulated patient data that are classified as dementia are obtained via a pre-trained classifier. The significant variables that cause the conversion from MCI to Alzheimer's disease are obtained via analysis of the change in the attributes of the MCI patients and the generated dementia patients from the MCI patients.
\item We investigate correlations between the change in variables that cause the label of a patient flipping from MCI to dementia.
\item We discover relationships between the principal directions in the latent space and the dementia diagnosis.
\item We show that the data of the synthetically generated dementia patients and the real MCI patients, who are diagnosed with dementia in the future time steps, share similar characteristics. This finding is a piece of evidence that the latent space manipulation applied in the proposed study may correspond to a clinically realistic transition between MCI and dementia. As a result, the proposed framework may reveal more information about a patient cohort than a deep classifier.
\end{itemize}
We conduct experiments with the dataset of The Alzheimer's Disease Prediction of Longitudinal Evolution, TADPOLE challenge~\cite{marinescu}. The credibility of the experimental results from a clinical perspective is confirmed with a neurologist.

\begin{figure*}[!t]
\centering
  \includegraphics[width=0.76\textwidth]{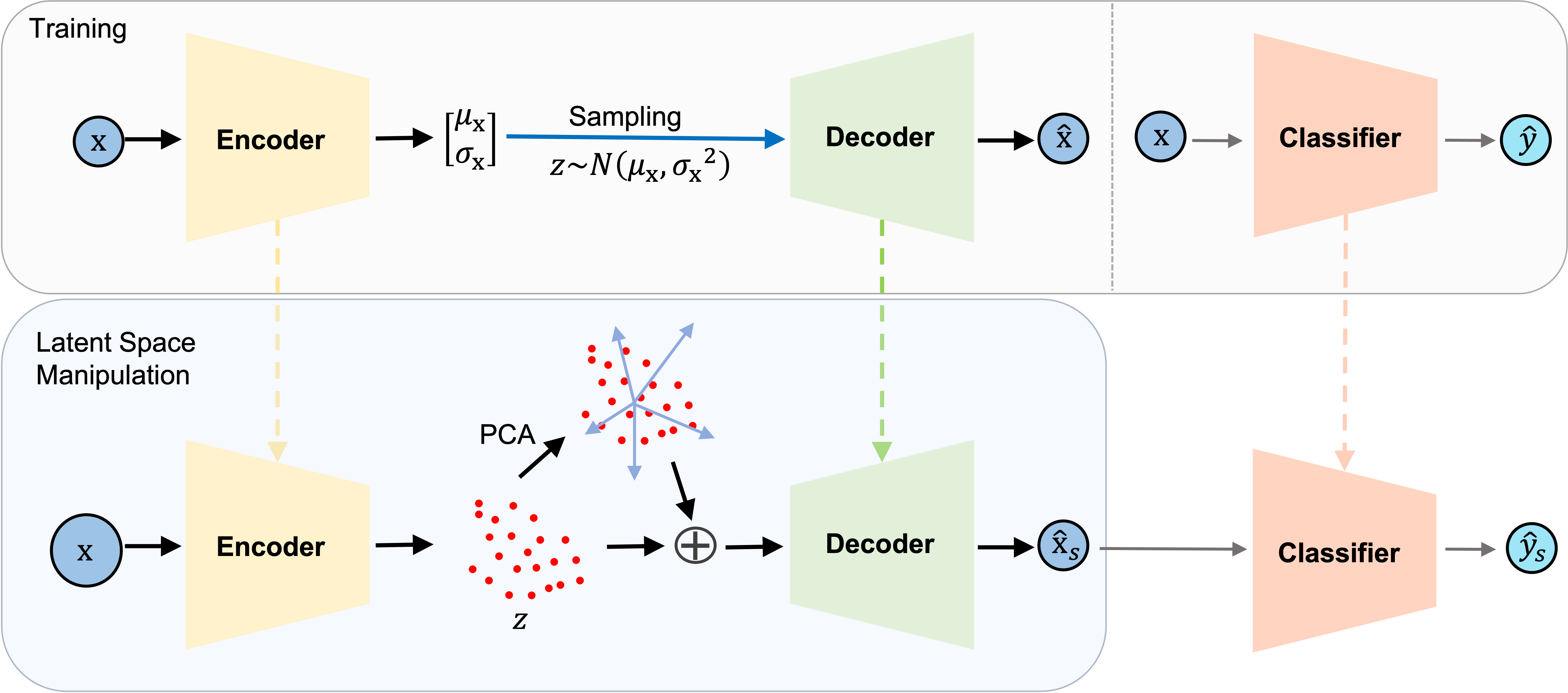}
  \caption{Proposed framework. A variational autoencoder and a binary classifier are trained using MCI and dementia patients. Principal components of the latent representations of the patients are computed. We empirically observe the first principal component as the direction of dementia in the latent space. Latent representation of an MCI patient is manipulated by adding the first principal component. Manipulated representations are decoded to obtain the synthetic patients denoted by $\hat{\mathbf{x}}_s$. Labels of the synthetic patients are verified using the binary classifier.} 
  \label{fig:model}
\end{figure*}

\section{Related Work}
\label{sect:related_work}
MCI is defined as a transitional stage between normal aging and dementia~\cite{rabin2009differential}. According to the clinical studies, not all MCI patients convert to AD~\cite{chapman2011predicting}. For this reason, detection of the MCI patients who will likely develop AD is of utmost importance for targeted clinical interventions. Therefore, there is a growing literature on the topic. Some studies solely focus on imaging techniques, while others consider the results of cognitive and physiological tests. In this section, we review clinical studies with traditional machine learning models as well as, the studies in the computer science domain which propose novel methods to solve the problem. 

\subsection{Clinical Studies on the Transition Between MCI and AD}
Biological and behavioral markers obtained through lab and neuropsychological tests are often used to detect AD~\cite{chapman2011predicting}. There has been a quest for finding noninvasive distinctive identifiers and the most informative way of fusing multiple identifiers to diagnose a patient with dementia. For instance, Lu {\it et al.} propose to fuse AD biomarkers with an event-based probabilistic framework, named CARE index~\cite{lu2020predicting}. The authors state that although screening abnormal amyloid and tau proteins has been essential for accurate diagnosis, it is not always possible~\cite{lu2020predicting}. For instance, amyloid-sensitive PET is required to measure the level of amyloid~\cite{teipel2020vivo}. Therefore, it is necessary to enable accurate and early diagnosis of AD considering the biomarkers obtained through noninvasive screening. In such cases, it is essential to analyze multiple factors simultaneously. For instance, Sun {\it et al.} consider a hierarchical Bayesian model to extract underlying factors from atrophy patterns and cognitive impairments simultaneously~\cite{sun}.

Mofrad {\it et al.} also focus on the abnormalities in brain atrophy due to aging~\cite{mofrad2021predictive}. The authors apply an ensemble of predictive models on MRI images to predict the conversion from cognitively normal patient to MCI and from MCI to AD~\cite{mofrad2021predictive}. Bae {\it et al.} also focus on the MRI data however they prefer a 3-dimensional Convolutional Neural Network (CNN) to extract features from structural MRI images~\cite{bae2021transfer}. When more complex models are utilized, an adequate amount of data is required. However, it is often not possible to attain large-scale patient data. Therefore, the authors resort to fine-tuning a pre-trained CNN~\cite{bae2021transfer}. Instead of directly utilizing the MRI images, some studies aim to discover new biomarkers from MRI. For instance, Kung {\it et al.} propose a biomarker, named ratio of principal curvatures (RPC), that may be an identifier to detect the transition between MCI and AD~\cite{kung2021neuroimage}. Kuang {\it et al.}, on the other hand, considered a drastically different dataset which is collected via a cognitive questionnaire. The authors work with less complex models, such as logistic regression, and Multi-layer Perceptron (MLP). Although the models used in the study may be more interpretable, relying only on the questionnaire dataset is a limitation~\cite{kuang2021prediction}. 

\subsection{Deep Learning for Neurodegenerative Diseases}
In computer science, various healthcare tasks, such as the progression of neurodegenerative diseases~\cite{termine2021multi,nguyen}, subtyping patients~\cite{baytas}, and early prediction of the conversion from MCI to AD, are usually posed as supervised and unsupervised learning problems with various challenges. Some of these challenges stem from the temporal, heterogeneous, multivariate, multi-view, and non-linear nature of the patient data. Therefore, traditional machine learning problems may not always be competent to tackle them. Consequently, deep learning techniques are preferred in healthcare due to their ability to capture complex interactions between data points.

Deep learning research for early diagnosis of AD comprises of two frequently encountered types of studies. There is a substantial amount of work regarding CNN architectures for feature extraction and segmentation from neuroimaging~\cite{basheera2020novel,amini2021diagnosis,wang2020explainable,parmar2020deep,abdulazeem2021cnn,xia2020novel}. These studies utilize effective deep frameworks to predict AD diagnosis. However, they do not address the revelation of significant factors behind the AD diagnosis due to the lack of interpretability. The second line of research is concerned with designing deep frameworks to learn predictive models from clinical, cognitive, neurophysiological test results, and genotype~\cite{venugopalan2021multimodal,bringas2020alzheimer,li2021use,zhou2021correlation}. In this category, some studies are aiming to extract interpretable features. For instance, Venugopalan {\it et al.} extract features from multiple modalities, such as, clinical, genetic, and imaging using a stacked denoising auto-encoder~\cite{venugopalan2021multimodal}. The authors address the interpretability challenge via a feature masking strategy where one feature is discarded at a time and a decrease in prediction accuracy is observed. According to the experiments, hippocampus, amygdala brain areas, and Rey Auditory Verbal Learning Test are identified as the most significant features for early detection of AD. Although Venugopalan {\it et al.}~\cite{venugopalan2021multimodal} points to a similar direction as this study, our proposed framework is concerned with deciphering the underlying reasons for the transition between MCI and AD. We model this transition using a latent space manipulation technique.

\subsection{Latent Space Manipulation}
Latent space manipulation is a relatively new concept that is usually applied to Generative Adversarial Networks (GANs). Controlling the latent space enables reshaping the generated output in certain ways. For example, one of the vision studies employs latent space manipulation to generate face images that look more feminine, older, or with more makeup~\cite{shen}. Another study from the healthcare domain utilizes latent space manipulation for high-resolution medical image synthesis using GANs~\cite{fetty2020latent}. Learning to control the latent space can be both during training or after training (by using a pre-trained model) as in this study. 

The latent space manipulation can be learned via supervised and unsupervised learning frameworks. In the supervised techniques, an attribute classifier is trained to discover relevant directions within the latent space~\cite{goetschalckx}. Principal Component Analysis (PCA) can also be applied to data points sampled from a latent space to find relevant directions~\cite{harkonen}. DeVries {\it et al.}, on the other hand, propose to apply simple transformations, such as extrapolation, interpolation, and adding Gaussian noise, to the latent representations learned via an auto-encoder~\cite{devries}. A more recent study proposes an automatic, modality-ignorant approach where the latent space is manipulated by extrapolation and interpolation~\cite{cheung2021modals}. The studies based on simple transformations aim data augmentation rather than generating very different samples than the input. 

In this study, we pursue unsupervised latent space manipulation with PCA. To the best of our knowledge, this is the first attempt to employ a latent space manipulation technique for investigating a healthcare problem using non-vision data.

\section{Method}
\label{sect:method}

\begin{algorithm}[!t]
\caption{Latent space manipulation for the MCI patients and synthesis of new patients.}
\begin{algorithmic}[1]
\Statex Obtain the latent representations of the patients:
\State $\mathbf{z}_i = \text{encoder}\left(\mathbf{x}_i\right)$, $i = 1, \cdots, N$
\Statex Compute the principal component:
\State $\mathbf{u} = \text{PCA}\left(\mathbf{z}_1, \cdots, \mathbf{z}_N\right)$
\Statex Manipulate the selected correctly classified MCI patients
\State $\mathbf{z}^{*}_{\text{MCI}} = \mathbf{z}_{\text{MCI}} + \alpha \mathbf{u}$
\Statex Reconstruct the manipulated latent representations
\State $\hat{\mathbf{x}}_s = \text{decoder}\left(\mathbf{z}^{*}_{\text{MCI}}\right)$
\end{algorithmic}
\label{alg:proposed}
\end{algorithm}

In this section, we present the proposed framework shown in Figure~\ref{fig:model}. The proposed model consists of a VAE and a binary classifier. Problem definition and the details of the proposed framework are presented in the following sections.

\subsection{Problem Definition}
Predicting the conversion from MCI to AD can be posed as a supervised learning problem. For this purpose, one would need a collection of MCI and AD patients. Thus, a binary classifier can be learned to distinguish the two categories. On the other hand, the goal of the proposed study is to investigate the reasons behind this progression. Therefore, the problem might also be posed as an unsupervised learning problem, where the patient characteristics are learned from the data. A classifier model is still learned to evaluate the outputs of the proposed unsupervised procedure. Throughout the study, we look for the answer to the following questions.
\begin{itemize}
    \item Which input attributes do play a significant role in the progression of the MCI to dementia?
    \item How do the significant attributes behave during the transition between MCI and dementia?
\end{itemize}

\subsection{Model Architecture}
The proposed framework require to train a VAE and a binary classifier before executing the latent space manipulation module. 

\subsubsection{Binary Classifier}
A binary classifier is trained to distinguish MCI and AD patients. The patient data might be temporal. However, in this study, we focus on each time step separately. Thus, an MLP network is learned on real MCI and AD patient data. The binary cross-entropy loss given below is used to train the classifier.
\begin{align}
\mathcal{L}_C = \frac{1}{N}\sum_{i=1}^{N}-\{y_i \log \hat{y}_i + \left(1-y_i\right)\log \left(1-\hat{y}_i\right)\}
    \label{eq:xent}
\end{align}
where $N$ denotes the number of training data points, $y_i$ is the groundtruth label of $i$th sample, $\hat{y}_i$ is the predicted label of the $i$th sample.

The trained classifier is used to verify whether the label of the synthetic patients who are generated after the latent space manipulation shifts from MCI to AD. The probability of the classifier also provides an insight into the manipulated patient and the amount of the manipulation towards the direction of AD. For instance, the high confidence of the classifier for a synthetic patient, who was labeled as AD, may mean that the condition of the manipulated MCI patient was close to being diagnosed with dementia in the future steps. 

\subsubsection{Variational Autoencoder}
The VAE network aims to learn a latent space for the input samples in an unsupervised way. Unlike the traditional autoencoders, the encoder network outputs a distribution rather than projecting the input to a fixed-sized latent vector. Thus, the VAE prevents overfitting to the specific patients in the training set but captures the distributions of MCI and AD patients. Since the decoder network is trained to reconstruct the input from a sample drawn from a normal distribution, a generative property is injected into it. These properties make the VAE's latent space more versatile for the latent space manipulation task than the traditional autoencoder. In this study, the encoder and decoder networks of the VAE are comprised of fully connected layers with ReLU activation functions, and they are trained to minimize the following loss function.
\begin{align}
    \mathcal{L}_{\text{VAE}} = \frac{1}{N}\sum_{i=1}^{N} ||\mathbf{x}_i - \hat{\mathbf{x}}_i||_{2}^{2} + \gamma \text{KL}\left[N\left(\boldsymbol\mu_{x}, \boldsymbol\sigma_{x}^{2}\right), N\left(0,1\right)\right]
    \label{eq:vae}
\end{align}
where $\mathbf{x}_i\in\mathbb{R}^{d}$ is the input, $\hat{\mathbf{x}}_i\in\mathbb{R}^{d}$ is the reconstructed input, $\boldsymbol\mu_x, \boldsymbol\sigma_{x}^{2}$ denote the mean and variance of the distribution learned by the encoder, $\text{KL}\left[\cdot\right]$ is the Kullback-Leibler divergence, and $\gamma$ controls the effect of KL divergence that measures the distance between two distributions.

\subsection{Latent Space Manipulation}
The latent space manipulation techniques are usually utilized with generative models for vision problems. There is a growing number of studies investigating interpretable directions to manipulate the latent space of a GAN to synthesize images~\cite{goetschalckx,shen,harkonen}. In this study, inspired by the vision domain, we propose to use latent space manipulation to investigate the factors that cause MCI patients to develop dementia. To the best of our knowledge, the latent space manipulation techniques were not applied to tabular patient data before. In the visual domain, the effects of the manipulation can be visually recognized on the synthesized images. On the other hand, in this study, we utilize the binary classifier to understand the effects of the manipulation.

Inspired by the latent space manipulation technique proposed by~\cite{harkonen}, important latent directions are identified by applying PCA to the latent representations of the patients. Then, the latent representation of a patient is manipulated and the decoder of the VAE is used to synthesize a new patient from the manipulated latent representation as in Equation~\ref{eq:manip}.
\begin{align}
\label{eq:manip}
\mathbf{z}^{*} &= \mathbf{z} + \alpha \mathbf{u} \\
\hat{\mathbf{x}}_{s} &= \text{decoder}\left(\mathbf{z}^{*}\right) \nonumber
\end{align}
where $\mathbf{z}$ is the latent representation learned by VAE, $\mathbf{u}$ is one of the principal components, $\alpha$ denotes the coefficient that controls the amount of manipulation, and $\hat{\mathbf{x}}_{s}$ denotes the synthesized patient. 

In this study, we aim to understand the underlying factors that cause a patient to be diagnosed with dementia. For this reason, we focus on manipulating the latent representations of MCI patients. For this purpose, first, the reconstructed MCI patient vectors are generated via the learned VAE. Then, we select the MCI patients whose reconstructed samples are correctly identified as MCI by the binary classifier for the latent space manipulation procedure. This procedure aims to select the correctly reconstructed data points so that the effect of manipulation on label shift can be clearly observed. After the latent representations of the selected MCI patients are manipulated using the Equation~\ref{eq:manip}, the predicted labels of the synthesized patients are obtained by the binary classifier to investigate the change that MCI patients undergo. The proposed procedure to synthesize new patients is also summarized in Algorithm \ref{alg:proposed}. Detailed analysis of the proposed approach is presented in the following section.

\begin{table*}[!t]
\centering
\small
\begin{tabular}{||l c c c||} 
 \hline
 \textbf{Variable} & \textbf{Dementia - mean (± std)} & \textbf{MCI - mean (± std)}& \textbf{Missing Value Percentage}  \\ 
 \hline\hline
  Clinical Dementia Rating Scale (SB) & 5.99 ± 3.07×$10^0$ & 1.62 ± 1.11×$10^0$& \%29.65 \\ 
 \hline
 RAVLT immediate & 2.03 ± 0.81 ×$10^1$ & 3.41 ± 1.12 ×$10^1$ & \%30.69 \\ 
 \hline
 RAVLT forgetting & 4.23 ± 1.93 ×$10^0$ & 4.66 ± 2.49×$10^0$& \%30.89 \\
 \hline
 RAVLT learning & 1.66 ± 1.72 ×$10^0$ & 3.99 ± 2.63 ×$10^0$& \%30.69  \\
 \hline
 ADAS-Cog11 & 2.29 ± 1.01 ×$10^1$ & 0.99 ± 0.51 ×$10^1$& \%30.06  \\
 \hline
  ADAS-Cog13 & 3.35 ± 1.13 ×$10^1$& 1.61 ± 0.71 ×$10^1$ & \%30.74  \\
 \hline
    Functional Activities Questionnaire & 1.69 ± 0.75 ×$10^1$& 0.34 ± 0.43 ×$10^1$ & \%29.41  \\
 \hline
    Mini-Mental State Examination & 2.12 ± 0.46 ×$10^1$&2.74 ± 0.22 ×$10^1$ & \%29.89  \\
 \hline
   Ventricles & 5.32 ± 2.54 ×$10^4$&4.19 ± 2.32 ×$10^4$ & \%41.57  \\
 \hline
   Hippocampus & 5.51 ± 1.11 ×$10^3$ & 6.71 ± 1.12 ×$10^3$& \%46.61  \\
 \hline
   Whole brain volume & 0.95 ± 0.11 ×$10^6$ & 1.02 ± 0.11 ×$10^6$ & \%39.65  \\
 \hline
     Entorhinal cortical volume & 2.69 ± 0.72 ×$10^3$ & 3.51 ± 0.77 ×$10^3$& \%49.23  \\
 \hline
     Fusiform cortical volume & 1.49 ± 0.26 ×$10^4$ & 1.74 ± 0.28 ×$10^4$& \%49.23  \\
 \hline
   Middle temporal cortical volume & 1.65 ± 0.32 ×$10^4$& 1.95 ± 0.29 ×$10^4$ & \%49.23  \\
 \hline
   Intracranial volume & 1.54 ± 0.18 ×$10^6$& 1.54 ± 0.16 ×$10^6$ & \%37.58  \\
 \hline
   APOE4 & 0.88 ± 0.71 ×$10^{0}$ & 0.55 ± 0.66 ×$10^{0}$& \%0.09  \\
 \hline
  Diagnosis & - & -& \%30.11  \\ 
 \hline
\end{tabular}
\caption{Set of variables with their statistics and percentage of missing values. SB: Sum of boxes, ADAS: Alzheimer’s Disease Assessment Scale, RAVLT: Rey Auditory Verbal Learning Test.}
\label{tab:variables}
\end{table*}

\section{Experiments}
\label{sect:exp}
The TADPOLE data used in the preparation of this study were obtained from the Alzheimer’s Disease Neuroimaging Initiative (ADNI) database (\url{adni.loni.usc.edu}). The ADNI was launched in 2003 as a public-private partnership led by Principal Investigator Michael W. Weiner, MD. The primary goal of ADNI has been to test whether serial MRI, PET, other biological markers, and clinical and neuropsychological assessment can be combined to measure the progression of MCI and early AD. For up-to-date information, see \url{www.adni-info.org}. Google Colab~\footnote{https://colab.research.google.com/notebooks/intro.ipynb} is used for training the proposed framework.

\subsection{Data Preprocessing}
TADPOLE data combines MRI, PET, CSF, diffusion tensor imaging (DTI), cognitive tests, some genetic and demographic information from the ADNI database~\cite{jack}. The dataset consists of visit information of $1737$ subjects who may be normal control (NC), MCI, or AD patients. Each row represents a patient visit, demonstrating the measurements and test scores as features. On average, each subject has $6.68$ different visits up to $6$ years from their first examination. The average number of years between the first and last time points is $3.5 \pm 2.5$. The TADPOLE challenge recommends $23$ variables including diagnosis, neuropsychological test scores, anatomical features derived from T1 MRI, PET, and CSF markers. However, we selected $16$ variables given in Table~\ref{tab:variables} which have less than $50\%$ of missing values.  

Missing data is imputed with forward filling method where the most recent measurement of the patient before the missing value is used for the missing value. If a variable cannot be imputed by the forward filling, the average value of the variable is calculated from the visits which have the same diagnostic label.
In the experiments, it is noticed that some measurements are anomalies such that their values are not clinically plausible. Such anomalies are removed from the dataset. PET and MRI measurements, namely hippocampus, whole brain, middle temporal cortical, entorhinal, and fusiform volumes are normalized by the intracranial volume of the particular patient. These measurements do not have standard values and vary according to the patient's head size. Therefore, they are standardized~\cite{sargolzaei,voevodskaya}.

Since this study analyzes the conversion from MCI to dementia, we eliminate the patients with only MCI and dementia labels. Thus, we have 1343 data points with dementia label and 4021 data points with MCI label. We use the standard scaling method for normalizing the numerical columns. Compared to min-max scaling, we empirically observe that the standard scaling performs better with the VAE.

\begin{figure}[!t]
\centering
  \includegraphics[width=0.35\textwidth]{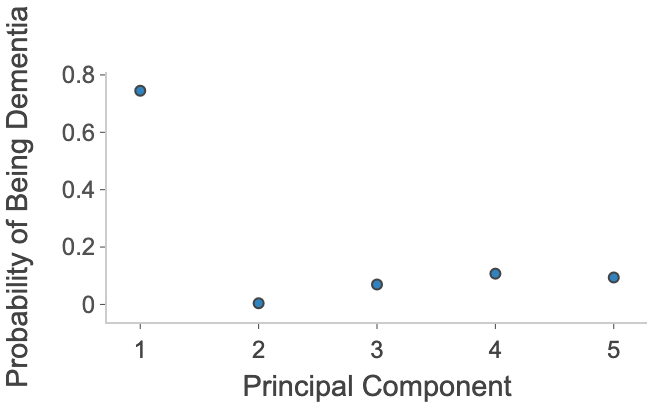}
  \caption{Manipulation effects of principal components for a fixed $\alpha$ value of $5$. The average probability of the classifier for dementia computed using the manipulated MCI patients is the highest for the first principal component.}
  \label{fig:pc_softmax}
\end{figure}

\begin{figure}[!t]
\centering
  \includegraphics[width=0.35\textwidth]{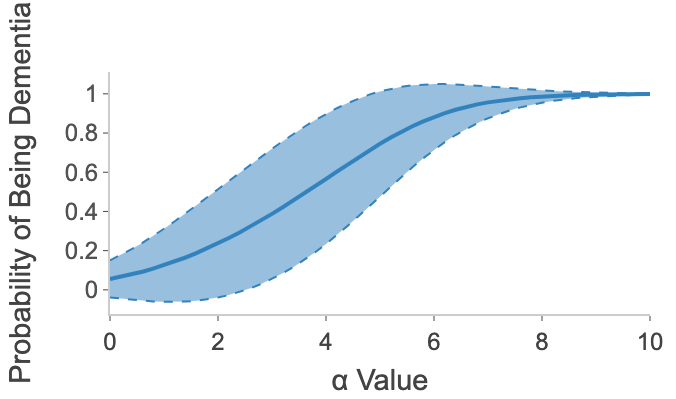}
     \caption{Manipulation effect of the first principal component with different $\alpha$ values. Horizontal axis represents $\alpha$ values, vertical axis represents the average probability of being dementia of the data points after manipulation and the shaded region denotes the standard deviation.}
    \label{fig:softmax_alpha}
\end{figure}

\begin{figure}[!t]
\centering
  \includegraphics[width=0.35\textwidth]{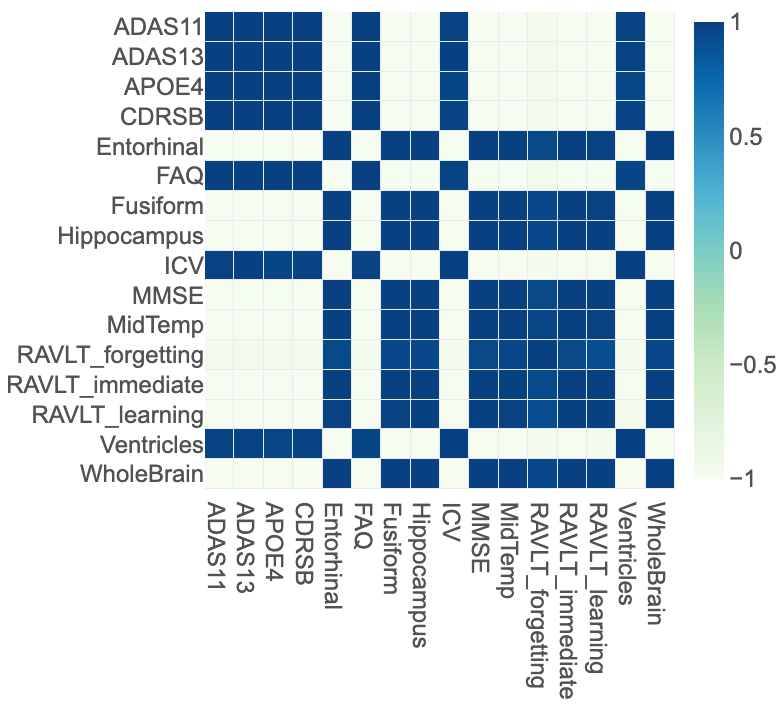}
     \caption{Correlations of the average changes in variables of the MCI patients who convert to dementia after manipulation with the first principal component with varying $\alpha$.}
     \label{fig:first_pc_heatmap}
\end{figure}



\begin{figure*}
     \centering
     \begin{subfigure}[b]{0.38\textwidth}
         \centering
         \includegraphics[width=\textwidth]{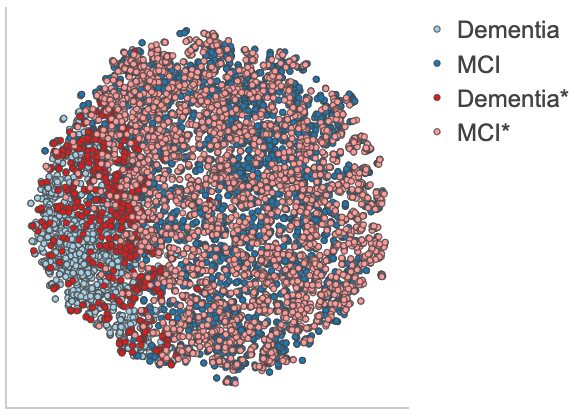}\caption{$\alpha=2$}\label{fig:a}
     \end{subfigure}
     \hfill
     \begin{subfigure}[b]{0.38\textwidth}
         \centering
         \includegraphics[width=\textwidth]{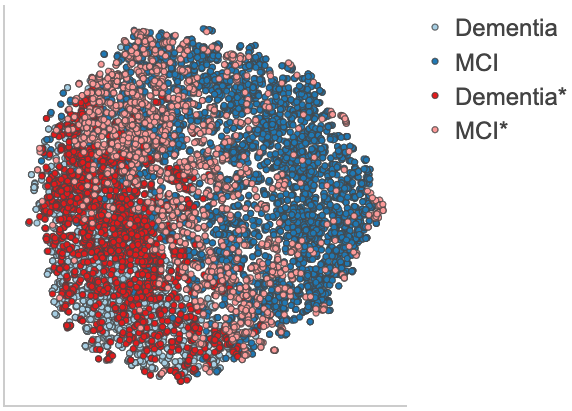}\caption{$\alpha=4$}\label{fig:b}
     \end{subfigure}
        \caption{t-SNE of the manipulated data points with their original versions in the classifier's latent space, with $\alpha=2$ and $4$. Light blue and dark blue colors represent Dementia and MCI data points before manipulation, red and pink colors represent Dementia and MCI data points after manipulation.}
       \label{fig:tsne_new}
\end{figure*}

\subsection{Training Scheme}
To train the VAE model, hyperparameter optimization is employed with the 5-fold cross-validation. The encoder is designed as three-layer fully connected network with the dimensions of $\left[256, 128, 64\right]$ for the first, second and third layer, respectively. Another hyperparameter to be tuned is $\gamma$ in Equation~\ref{eq:vae}. Higher values provide more structured latent space with weaker reconstruction, and lower values provide better reconstruction with less structured latent space~\cite{higgins}. The parameter $\gamma$ is set to $0.1$ after evaluating the results for the hyperparameter space of $\left[1,0.75,0.5,0.25, 0.1\right]$.

The binary classifier is designed as a two-layer fully connected network. Hidden dimensionality is determined as $128$ and $64$ by 5-fold cross-validation. Batch normalization and dropout with a probability of $0.2$ are used for regularization. Both the binary classifier and variational autoencoder models are trained with batch size of $128$ and validation set split ratio of $0.1$. Early stopping is used as regularization with $100$ epochs. Adam optimizer is used with a learning rate scheduler where the decay rate is $0.1$ and the initial learning rate is set to $0.001$. During training and data preprocessing, the random seed is fixed. The classification performance on test data is $85\%$ of ROC, $92\%$ of precision and $96\%$ of recall. 

\begin{table*}[!t]
\centering
\small
\begin{tabular}{||l c c ||} 
 \hline
 \textbf{Variable} & \textbf{Manipulated Data vs. Original Data} & \textbf{Future Visit Data vs. Original Data}  \\ 
 \hline\hline
  Clinical Dementia Rating Scale (CDRSB) & \%27.85 & \%46.29 \\ 
 \hline
 RAVLT immediate & \%-8.01 & \%-7.31  \\ 
 \hline
 RAVLT forgetting & \%-0.56 & \%-2.15  \\
 \hline
 RAVLT learning & \%-21.01 & \%-15.52   \\
 \hline
 ADAS-Cog11 & \%15.43 & \%15.49   \\
 \hline
  ADAS-Cog13 & \%13.23 & \%13.24  \\
 \hline
    Functional Activities Questionnaire (FAQ) & \%28.38 & \%47.88   \\
 \hline
    Mini-Mental State Examination & \%-4.4 & \%-6.49  \\
 \hline
   Ventricles & \%5.79 & \%6.87   \\
 \hline
   Hippocampus & \%-6.67 & \%-4.93  \\
 \hline
   Whole brain volume & \%-2.25 & \%-2.39  \\
 \hline
     Entorhinal cortical volume & \%-6.84 & \%-4.98   \\
 \hline
     Fusiform cortical volume & \%-3.77 & \%-2.68   \\
 \hline
   Middle temporal cortical volume & \%-3.76 & \%-3.04   \\
 \hline
   Intracranial volume (ICV) & \%0.99 & \%-0.1   \\
 \hline
   APOE4 & \%13.36 & \%9.59  \\ 
 \hline
\end{tabular}
\caption{Average changes between the manipulated and original time steps that are labeled as dementia for a fixed $\alpha=1$. The manipulation results in similar changes as the real life for the majority of the variables. The manipulation effect falls behind on changing CDRSB and FAQ variables accordingly, and misinterprets the effect of ICV. }
\label{tab:change_perc}
\end{table*}

\subsection{Results and Discussion}
The latent space manipulation experiments are conducted with varying values of $\alpha \in \left[0.2, 0.4,\cdots, 10\right]$. As given in Equation~\ref{eq:manip}, $\alpha$ is the coefficient of the principal components (PCs). We focus only on positive values of $\alpha$ since we would like to manipulate the latent representation towards the direction of the PC. 

\subsubsection{Choice of Principal Components}
We extract the top $5$ PCs from the patient's latent representations using PCA. Each PC is added to the latent representations of the MCI patients as in Equation~\ref{eq:manip}. The manipulation effect of each PC is evaluated by the number of MCI patients who are classified as dementia after the manipulation. In Figure~\ref{fig:pc_softmax}, for a fixed value of $\alpha=5$, we plot the average probability of the classifier for dementia computed using manipulated MCI patients against the $5$ different principal directions. This empirical evidence indicates that the first PC, which captures the most informative direction, moves the latent representation of an MCI patient towards dementia. For the rest of the experiments, the first PC is considered. 
\subsubsection{Investigating the Effect of \textbf{$\alpha$}}
Next, we investigate the effect of varying $\alpha$ values. Figure~\ref{fig:softmax_alpha} demonstrates the average of softmax probabilities of being dementia against different $\alpha$ values for the first principal component. Different $\alpha$ values have a non-linear behavior for converting the MCI patients to dementia. We observe similar behavior when the number of manipulated MCI patients labeled as dementia is investigated for each $\alpha$ value.  As we show in Figure~\ref{fig:softmax_alpha}, when the $\alpha$ value reaches $5$, we obtain the highest incremental manipulation effect. 
\subsubsection{Clinical Interpretation}
As the manipulation effect increases, values of some variables increase and some of them decrease. We first analyze positive and negative relationships between the change in values of the variables when we increase $\alpha$. For this purpose, we focus on the MCI patients who convert to dementia after the manipulation. Correlation between the average change in variables after the manipulation with the first PC for increasing $\alpha$ is shown in Figure~\ref{fig:first_pc_heatmap}. In the figure, dark-colored cells indicate a positive correlation while the light color cells indicate an opposite behavior. For instance, the values of ADAS-11, ADAS-13, APOE4, FAQ, ICV, and Ventricles increase together while the whole brain volume behaves opposite to them. Another example could be the positive relationship between MMSE and RAVLT measurements. As we move towards dementia, MMSE and RAVLT measurements decrease together. In Figure~\ref{fig:first_pc_heatmap}, we do not show if the variables increase or decrease after manipulation. However, this information can be retrieved from Table~\ref{tab:change_perc}.

To investigate the individual changes in variables after manipulation, we manipulated the MCI patients, who are diagnosed with dementia in future time steps, using the first PC and $\alpha$ of $1$. We compare the average changes between the manipulated and original time steps that are labeled as dementia in Table~\ref{tab:change_perc}. Future visit information after the manipulated data points of the corresponding patients shows that the patients undergo similar changes to be diagnosed with dementia. The results are clinically plausible for the dataset and the variables used in this study. The reported results in this section imply that we can discover a direction that contains a significant amount of information about AD in the latent space of a deep network.

\subsubsection{Visualizing Manipulated Data in Classifier's Latent Space}
Figure \ref{fig:tsne_new} visualizes manipulated data points with their original counterparts in the latent space of the binary classifier. Since we apply manipulation only on the MCI labeled patients, Dementia$^*$ and MCI$^*$ represent manipulated versions of MCI data points. With this visualization, we can observe the diversity in generated data points. As expected, some of the generated dementia patients are close to the boundary of MCI and dementia regions. However, as desired, they are not collapsed in the middle especially for higher $\alpha$ as in Figure~\ref{fig:b}. We infer from this experiment that the proposed approach may be suitable for generating realistic synthetic dementia patients under the low data regime.

\section{Conclusion}
Detection of the conversion from MCI to AD is essential for timely treatments and decelerate the disease progression. In this study, we propose a deep framework to investigate the factors that cause this conversion. For this purpose, a VAE network and a binary classifier are trained using the TADPOLE challenge dataset~\cite{marinescu}. We discover that the first principal component computed from the latent representations of MCI and dementia patient visit data provides the direction towards the conversion to dementia. Thus, we manipulate the latent representations of the MCI patients using the first principal component and obtain dementia patients. The experimental results indicate that it is possible to discover a clinically meaningful direction in the latent space of a deep network. According to the promising results, the proposed framework shows potential to reveal underlying information captured by deep models and use this information to generate realistic synthetic patient data.

\section{Acknowledgments}
This work is supported by Boğaziçi University Research Fund under grant number 17004. This work uses the TADPOLE data sets \url{https://tadpole.grand-challenge.org} constructed by the EuroPOND consortium \url{http://europond.eu} funded by the European Union’s Horizon 2020 research and innovation program under grant agreement No 666992. TADPOLE Data was obtained from Alzheimer’s Disease Neuroimaging Initiative (ADNI). Data collection and sharing for this project was funded by the Alzheimer’s Disease Neuroimaging Initiative (ADNI) (National Institutes of Health Grant U01 AG024904) and DOD ADNI (Department of Defense award number W81XWH-12-2-0012). ADNI is funded by the National Institute on Aging, the National Institute of Biomedical Imaging and Bioengineering, and through generous contributions from the following: AbbVie, Alzheimer’s Association; Alzheimer’s Drug Discovery Foundation; Araclon Biotech; BioClinica, Inc.; Biogen; Bristol-Myers Squibb Company; CereSpir, Inc.; Cogstate; Eisai Inc.; Elan Pharmaceuticals, Inc.; Eli Lilly and Company; EuroImmun; F. Hoffmann-La Roche Ltd and its affiliated company Genentech, Inc.; Fujirebio; GE Healthcare; IXICO Ltd.; Janssen Alzheimer Immunotherapy Research \& Development, LLC.; Johnson \& Johnson Pharmaceutical Research \& Development LLC.; Lumosity; Lundbeck; Merck \& Co., Inc.; Meso Scale Diagnostics, LLC.;  NeuroRx Research; Neurotrack Technologies; Novartis Pharmaceuticals Corporation; Pfizer Inc.; Piramal Imaging; Servier; Takeda Pharmaceutical Company; and Transition Therapeutics. The Canadian Institutes of Health Research is providing funds to support ADNI clinical sites in Canada. Private sector contributions are facilitated by the Foundation for the National Institutes of Health (www.fnih.org). The grantee organization is the Northern California Institute for Research and Education, and the study is coordinated by the Alzheimer’s Therapeutic Research Institute at the University of Southern California. ADNI data are disseminated by the Laboratory for Neuro Imaging at the University of Southern California.
\newpage
\bibliography{aaai22} 
\end{document}